О.В. Палагін, академік НАН України
М.Г. Петренко, с.н.с., к.т.н.,
В.Ю. Величко, с.н.с., к.т.н.,
К.С. Малахов, магістр
О.В. Карун, магістр

Інститут кібернетики імені В.М.Глушкова Національної академії наук України

*В роботі описані основи методології проектування знання-орієнтованих інформаційних систем на основі онтологічного підходу. Такі системи реалізують технологію добування предметно-орієнтованих знань із множини природномовних текстів, їх формально-логічного представлення та прикладної обробки.*

# Основи проектування та розробки програмних моделей онтолого-керованих комп'ютерних систем

Для напряму онтологічного інжинірингу, стану й розвитку знання-орієнтованих комп'ютерних систем та пов'язаних з ними методів комп'ютерної обробки природномовної інформації (ПМ-інформації) і предметних знань існує розрив між добре опрацьованими методами та засобами для окремих етапів обробки ПМ-інформації для вирішення прикладних завдань у вузькоспеціалізованих предметних областях, з одного боку, і недостатністю таких для вирішення комплексних завдань, пов'язаних з аналізом і розумінням ПМ-інформації, її формально-логічним поданням, добуванням предметних знань з їх подальшим використанням у довільних (у науково-технічній сфері) предметних областях (ПдО).

Звідси випливає актуальність і важливість проблеми вироблення нових наукових знань, що включають методи й підходи до автоматичного аналізу і глибинно-семантичному розумінню ПМ-інформації, її формалізованого представлення в рамках відповідної теорії, автоматичного виявлення і вилучення нових знань, відповідних технологій та інструментальних засобів автоматизованої побудови онтологічних баз знань предметних областей [1,2].

Методологія проектування *онтолого-керованих комп'ютерних систем* (ОККС) представляє собою інтегровану сукупність методів, механізмів, моделей, алгоритмів та засобів для:

- ефективної обробки предметно-орієнтованих знань (ПОЗ), структурованих в онтологічні описи об'єктів, процесів і задач із заданої ПдО;
- автоматизованого добування знань з великих об'ємів предметно-орієнтованих природномовних текстів та побудови *комп'ютерних онтологій* (КО). В свою чергу, КО є не тільки інформаційною структурою ПОЗ, а й інструментальним засобом їх обробки;
- проектування архітектурної й інформаційної складових ОККС та алгоритмів вирішення задач користувачів;
- подальшого розвитку методології системної інтеграції ПОЗ та вирішення нагальних проблем міждисциплінарних наукових досліджень, зокрема побудову "електронних" наукових баз знань [3].

Методологія складається з: 1) методів системно-онтологічного аналізу заданої проблемної області;

2) методів і механізмів онтологічного підходу до проектування ОККС; моделей та алгоритмів обробки ПОЗ та побудови інструментального комплексу онтологічного призначення [3,4].

90-ті роки минулого сторіччя вважаються початком зародження *парадигми комп'ютерних онтологій* (КО). Вона була сформована як спроба усунути різного роду протиріччя, що все частіше проявлялися при функціонуванні та впровадженні інтелектуальних систем з використанням баз знань (БЗ) предметних областей. *Основними принципами* побудови КО є: дохідливість, ясність; обґрунтованість, зв'язність; розширюваність; мінімальний вплив кодування; мінімальні онтологічні зобов'язання [2].

***Комп'ютерна онтологія ПдО*** – це: 1) ієрархічна структура кінцевої множини понять, що описують задану ПдО; 2) структура представляє собою онтограф, вершинами якого є поняття, а дугами – семантичні відношення між ними; 3) поняття та відношення інтерпретуються відповідно із загальновизнаними функціями інтерпретації, взятими із електронних джерел знань із заданої ПдО; 4) додаткові інтерпретації понять та відношень визначаються аксіомами та обмеженнями їх області дії; 5) формально онтограф описується на одній з мов опису онтологій; 6) функції інтерпретації й аксіоми описані в деякій підходящій формальній теорії.

Схема формальної моделі онтології описується четвіркою [4]:

$$O = <X, R, F, A(D, Rs)>,$$

де $X$ – множина концептів; $R$ – множина концептуальних відношень між ними; $F: X \times R$ – кінцева множина функцій інтерпретації; $A$ – кінцева множина аксіом, які використовуються для запису завжди істинних висловлювань (визначень $D$ та обмежень $Rs$).

Парадигма КО, що розвивається у взаємодії із засобами та методами системного аналізу, поклала початок розвитку нової гілки засобів і методів системного аналізу ПдО – системно-онтологічного аналізу (або підходу). Його *центральною ідеєю є* розробка онтологічних засобів підтримки вирішення прикладних задач – *поліфункціональної онтологічної системи*. Така система описується кортежем, що складається з онтологій об'єктів, процесів та задач [4]:

$$ОнС = \langle O^{ПдО}(O^О, O^П), O^З \rangle.$$

*Онтологічний підхід* (ОнП) до проектування інформаційної й архітектурної компонент ОККС виник як міждисциплінарний підхід до побудови, представлення та обробки ПОЗ, моделі яких описують структуру та взаємозв'язки об'єктів відповідних ПдО.

ОККС з обробкою знань, що містяться в природномовних об'єктах, спроектована з врахуванням *онтологічного підходу*, якщо вона має наступні характерні риси [2]: 1) КО забезпечують ефективну машинну обробку мовних та предметних знань; 2) системно-онтологічний підхід допускає строгу систематизацію знань будь-якого рівня, в тому числі категоріального. Останній представляється *онтологією верхнього рівня*. Її проектування входить до загального алгоритму синтезу ОККС; 3) архітектурно-технологічні особливості ОККС: а) онтолого-керована архітектура характеризується високим рівнем формалізації представлення онтології ПдО та ефективними механізмами онтолого-керування, в тому числі підтримуючими розвиток системи; б) високий ступінь інтеграції міждисциплінарних знань; 4) використання засобів підтримки автоматизованої побудови онтологій

ПдО (методика, технологія та програмно-апаратна реалізація); 5) прикладна направленість та сильний взаємозв'язок технології обробки інформації з архітектурно-структурною організацією ОККС; 6) проектування ОККС виконується на основі принципів, методів та механізмів ОП; 7) функціонування ОККС в двох режимах: накопичення предметно-орієнтованих онтологічних знань та їх обробки.

### Основи інфологічного підходу проектування ОККС.

Такий підхід враховує логіко-інформаційну та онтологічну концепції проектування, а також віртуальну парадигму, за якої архітектура комп'ютерної системи орієнтована на технологію реконфігуровного процесинга. Остання забезпечує адаптивність системи завдяки наявності в ОККС архітектурних та технологічних можливостей настройки в умовах апріорної та поточної невизначеностей на основі навчання й досвіду.

Проектування ОККС допускає розробку двох взаємозалежних підсистем, відповідно для обробки знань в заданій ПдО й обробки текстів на основі "мовних" знань. Вказані підсистеми представлені відповідно як онтолого-керована КС обробки предметних знань (ОККС ПдО) та мовно-онтологічна КС обробки ПМ-інформації на основі мовних знань. Взаємодія між ними здійснюється шляхом реалізації процедури деякого напівформального відображення $G$.

*Модель ОККС*, що проектується, для вирішення типового набору задач в заданій проблемній області представлено системою $S = <M, A, X, P>$, де: $M$ – множина математичних методів; $A$ – множина алгоритмів, що реалізують множину методів $M$; $X$ – множина об'єктів і процесів, що описують задану ПдО і беруть участь в реалізації алгоритмів $\{A_i\}$ вирішення $m$-го типового набору задач; $P$ – процедура онтологічного опису об'єктів, процесів і задач заданої ПдО.

КС з онтолого-керованою архітектурою притаманні наступні характерні риси: композиція онтологій різного рівня й призначення, як по вертикалі, так і по горизонталі; ефективне багаторазове використання онтології ПдО й онтології задач для різних наборів типових задач; наслідком композиції принципів і механізмів подвійної парадигми предметних знань й онтологічного керування є інтеграція та ефективне використання компонентів архітектури ОККС в архітектурно-структурну організацію інструментального комплексу онтологічного призначення автоматизованої побудови онтологічних БЗ ПдО.

Блок-схема ОККС ПдО представлена на рисунку 1.

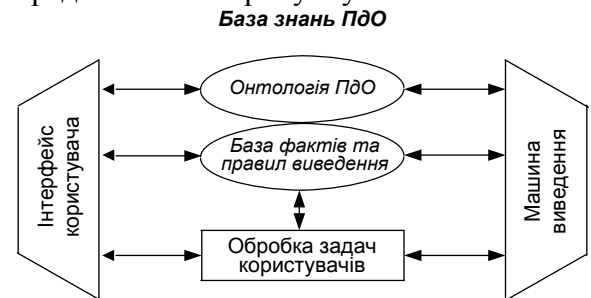

Рис. 1. Блок-схема ОККС ПдО.

### *Задачі та особливості проектування ОККС*

Процес проектування ОККС представляється у вигляді послідовності етапів (в основному системного, алгоритмічного та логічного), на кожному із яких проект представлений сукупністю математичних моделей, що описують різні її частини. Вказана сукупність математичних моделей тісно пов'язана з системою взаємозв'язаних алгоритмів, які, у свою чергу, описують відповідну множину вирішуваних задач і в своїй сукупності представляють *загальний алгоритм проектування ОККС*.

1. *Постановка задачі:* дослідження заданої ПдО; аналіз класу вирішуваних задач в заданій ПдО; вибір основних критеріїв проектування ОККС.
2. *Розробка онтолого-інфологічної моделі ОККС:* розробка інформаційної моделі ОККС; розробка онтологічної моделі мовних знань; розробка онтологічної моделі предметних знань (онтології ПдО); розробка моделі системної інтеграції мовних і предметних знань; розробка моделі формально-логічного опису мовних і предметних знань [1,5,6].
3. *Розробка системи взаємозалежних алгоритмів функціонування ОККС:* розробка алгоритмів функціонування мовно-онтологічної комп'ютерної системи; розробка алгоритмів функціонування ОККС ПдО; розробка алгоритмів переходу від обробки ПМ-інформації до обробки предметних знань.
4. *Розробка архітектури та структури ОККС:* розробка архітектурно-структурної організації мовно-онтологічної комп'ютерної системи (розробка знання-орієнтованого лінгвістичного процесора, розробка бази знань лексики ПМ); розробка архітектурно-структурної організації ОККС ПдО; розробка інтерфейсу користувача.
5. *Перевірка функціонування ОККС відповідно з заданими критеріями проектування.*

Прикладом розробки онтолого-керованої комп'ютерної системи за наведеним загальним алгоритмом проектування може бути інструментальний комплекс онтологічного призначення (ІКОН), головна функція якого полягає у формуванні онтологічних баз знань із довільних предметних областей (рис. 2).

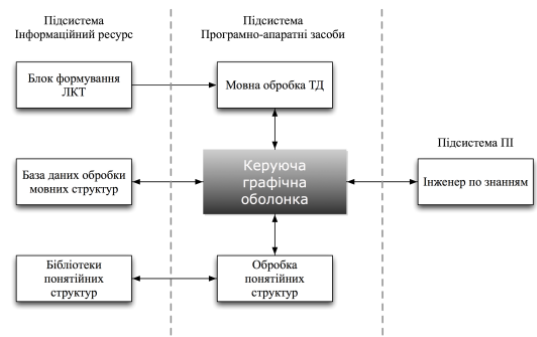

Рис. 2. Загальна блок-схема ІКОН.

ІКОН призначений для реалізації ряду інформаційних технологій:
− пошук у мережі Internet і/або в інших електронних колекціях текстових документів (ТД), релевантних заданої Пдо, їхню індексацію й збереження в базі даних;
− автоматична обробка природно-мовних текстів (ПМТ) (Natural Language Processing);
− добування з множини ТД знань, релевантних заданої ПдО, їх системно-онтологічну структуризацію й формально-логічне подання на одному (або декількох) із загальноприйнятих мов опису онтологій (Knowledge Representation).
− системна інтеграція онтологічних знань як один з основних компонентів методології міждисциплінарних наукових досліджень.

Програмна реалізація ІКОН повинна відповідати наступним вимогам [7]:
− наявність платформонезалежної керуючої графічної оболонки (КГО);
− модульна архітектура системи;
− використання загально-доступних засобів розробки програмного забезпечення;
− автоматичне конструювання (зв'язування) програмних модулів для вирішення цільової задачі;
− функціонування та інтеграція програмних модулів у розподіленому обчислювальному середовищі.

Наведеним вимогам відповідають системи з відкритою архітектурою на основі семирівневої моделі протоколів взаємодії відкритих систем OSI (Open System Interconnect).

Підсистеми ІКОН, взаємодіючи між собою, реалізують сукупність алгоритмів автоматизованої ітераційної побудови понятійних структур предметних знань, їхнього накопичення й системної інтеграції.

До підсистеми інформаційних ресурсів відносяться: електронні колекції енциклопедичних і тлумачних словників; мережа Internet; джерела онтологій предметних областей; бази даних ТД; база списків множин термінів, понять і відношень; пам'ять графового подання ТД. До підсистеми програмно-апаратних засобів відносятья: КГО ІКОН, інструментальна система Protege, когнітивний лінгвістичний процесор, пошукова система, програмні модулі керування бібліотеками, візуального проектування, формування формалізованого опису онтології, формування множини термінів, понять, та відношень, побудови онтології ТД та ПдО; модуль системної інтеграції онтологій.

Прикладний рівень (Application layer) у моделі OSI ідентифікує та встановлює наявність модулів для зв'язку, синхронізує спільно працюючі прикладні процеси, а також встановлює й узгоджує процедури усунення помилок і здійснює керування цілісністю інформації. Представницький рівень (Presentation layer) відповідає за трансформацію та узгодження форматів подання фактичних даних користувача, а також структур даних, які використовують програми. Функціонування прикладного і представницького рівнів забезпечує КГО ІККС ІКОН. КГО має стандартизований інтерфейс для підключення окремих функціональних модулів. В якості формату для обміну даними між окремими модулями у системі використовується мова XML. Сеансовий (Session layer) рівень також забезпечується КГО за рахунок підтримки взаємодії окремих модулів за протоколами PPTP (Point-to-Point Tunneling Protocol) та RPC (Remote Procedure Call). На нижчих рівнях використовуються стандартні протоколи, підтримка яких забезпечується за допомогою функцій різних операційних систем.

КГО інтегрує всі складові ІККС ІКОН в одне об'єднане середовище й виконує наступні функції [7]:

– у взаємодії з інженером по знаннях здійснює попереднє наповнення середовища зовнішніми електронними колекціями енциклопедичних, тлумачних словників і тезаурусів, що описують домен предметних знань;
– забезпечує запуск і послідовність виконання програмних модулів, що реалізують окремі інформаційні технології проектування онтології ПдО та системної інтеграції міждисциплінарних знань. Окремим випадком є автоматизована побудова тезаурусів ПдО для пошукових систем;
– відображає хід процесу проектування, ініціює повідомлення про поточний стан проекту та його наповнення інформаційними ресурсами;
– забезпечує обмін інформацією між програмними модулями і базами даних через загальну інформаційну шину.

Створення методології проектування *ОККС*, розробка програмних моделей і відповідних *інтегрованих інформаційних технологій* забезпечать виконання науково-дослідних робіт не тільки в конкретній предметній галузі, а і впродовж вирішення складних міждисциплінарних наукових досліджень.